\documentclass[conference]{IEEEtran}
\usepackage{amsmath,amssymb,amsfonts}
\usepackage{graphicx}
\usepackage{cite}
\usepackage{url}
\usepackage{authblk}

\title{A Graph Attention Network-Based Framework for Reconstructing Missing LiDAR Beams}

\author[1,*]{Khalfalla Awedat}
\author[2]{Mohamed Abidalrekab}
\author[3]{Mohammad El-Yabroudi}

\affil[1]{Computer Information Technology Department, SUNY Morrisville, 
Morrisville, NY, USA}
\affil[2]{Electrical and Computer Engineering Department, 
Portland State University, Portland, OR, USA}
\affil[3] {Electrical and Computer Engineering Department, Lawrence Technological University, Southfiled, MI, USA}
\affil[*]{\textit{awedatk@morrisville.edu}}

\begin{document}

\maketitle

\begin{abstract}
Vertical beam dropout in spinning LiDAR sensors – triggered by hardware aging, dust, snow, fog, or bright reflections – removes entire vertical slices from the point cloud and severely degrades 3D perception in autonomous vehicles. This paper proposes a Graph Attention Network (GAT)-based framework that reconstructs these missing vertical channels using only the current LiDAR frame, with no camera images or temporal information required.
Each LiDAR sweep is represented as an unstructured spatial graph: points are nodes and edges connect nearby points while preserving the original beam-index ordering. A multi-layer GAT learns adaptive attention weights over local geometric neighborhoods and directly regresses the missing elevation (z) values at dropout locations.
Trained and evaluated on 1,065 raw KITTI sequences with simulated channel dropout, the method achieves an average height RMSE of 11.67 cm, with 87.98\% of reconstructed points falling within a 10 cm error threshold. Inference takes 14.65 seconds per frame on a single GPU, and reconstruction quality remains stable for different neighborhood sizes k. These results show that a pure graph-attention model operating solely on raw point-cloud geometry can effectively recover dropped vertical beams under realistic sensor degradation.
\end{abstract}
\section{Introduction}

LiDAR sensing has become fundamental to modern autonomous driving systems, providing dense 3D measurements that support obstacle detection, ground estimation, and overall scene understanding \cite{ref9,ref10}. Rotating multi-beam LiDAR units generate a structured point cloud through a fixed set of vertical channels, each contributing a scan line as the sensor sweeps the environment. The integrity of these vertical beams is critical: they encode the elevation structure of the scene and determine how reliably the system can distinguish drivable surfaces from obstacles.

In practice, however, LiDAR acquisitions are often incomplete. Vertical beam dropout—caused by hardware aging, calibration drift, reflective materials, fog, dust, or snow—removes entire slices of the scan and produces discontinuities in the elevation profile \cite{ref11}. Several studies have documented how missing or corrupted beams degrade downstream perception, including object detection, depth reasoning, and free-space estimation \cite{ref12,ref13}. The impact is especially pronounced in the vertical dimension, where height information ($z$ or $r\sin\phi$) plays a central role in maintaining geometric consistency and identifying obstacles around the vehicle \cite{ref14,ref15}.

Most existing LiDAR-processing pipelines are designed to operate on whatever points the sensor returns, focusing primarily on classification or segmentation rather than repairing missing data. Approaches that attempt to fill gaps typically rely on voxel-based CNNs or interpolation methods, which struggle to capture the irregular, non-uniform structure of real scans and often overlook the relationships between adjacent beams \cite{ref16}. At the same time, recent developments in graph neural networks (GNNs) have shown strong performance on LiDAR tasks that benefit from modeling spatial neighborhoods and beam-index relationships \cite{ref3,ref17,ref18}. Within this family, Graph Attention Networks (GATs) \cite{ref1} are particularly well suited because they adaptively weight contributions from neighboring points, allowing the model to emphasize the most informative geometric cues.

This work addresses the specific challenge of reconstructing missing vertical beams in sparse LiDAR scans. We represent each LiDAR sweep as an unstructured spatial graph in which points are nodes and edges connect local geometric neighbors. Instead of relying on camera images or temporal aggregation, the problem is formulated entirely within a single LiDAR frame. The goal is to recover the missing elevation values at dropout locations by leveraging the structural continuity that exists across adjacent beams and along each scan line.

To accomplish this, we develop a multi-layer GAT framework tailored to beam reconstruction. By aggregating information through attention-based message passing, the network learns how height varies across local neighborhoods and how this variation reflects the underlying geometry of the scene. This enables the model to restore the vertical structure in a manner that preserves both fine-grained detail and global consistency. The proposed formulation offers a purely LiDAR-based solution capable of reconstructing vertical dropout in environments where camera data is unavailable or unreliable.

The remainder of this paper is organized as follows. Section~II reviews related work on LiDAR degradation, reconstruction, and graph-based processing. Section~III presents the proposed graph formulation and GAT reconstruction method. Section~IV reports experimental results on raw KITTI sequences, and Section~V concludes the paper.

\section{Related Work}

Early studies on LiDAR perception have shown that the spatial structure of point clouds is highly sensitive to sparsity and vertical sampling density. The survey in \cite{ref9} highlights how reduced vertical resolution directly weakens 3D object localization, particularly for tall or partially occluded objects. Similarly, the authors in \cite{ref13} demonstrate that changes in beam distribution measurably affect vehicle detection performance, reinforcing the importance of maintaining consistent vertical coverage. The work in \cite{ref14} further shows that the vertical arrangement of points carries strong geometric cues used in modern detectors, while \cite{ref15} links vertical geometry to ground–object separation and obstacle recognition. Together, these studies make clear that missing vertical beams can degrade downstream perception reliability, especially in autonomous driving scenarios.

Several approaches have attempted to compensate for sparse LiDAR data by generating denser or more structured point clouds. The GLiDR model introduced in \cite{ref3} employs a topologically regularized graph generative network to recover missing structure, but its objective focuses on global shape consistency rather than restoring beam-level information. The method in \cite{ref4} enhances 3D detection under limited LiDAR points through learned depth cues and spatial priors, yet it does not explicitly reconstruct dropped LiDAR channels. While these techniques reduce the impact of sparsity, they do not address the specific challenge of recovering missing vertical measurements caused by hardware failures or environmental degradation.

\section{Proposed GAT Architecture for Vertical Beam Recovery}

Graph-based representations have also gained traction in LiDAR processing. The survey in \cite{ref17} provides a broad overview of graph neural networks for 3D point clouds, emphasizing their ability to model local geometric relationships through flexible neighborhood structures. GAT, introduced in \cite{ref1}, further extend this idea by allowing nodes to weight neighbors based on feature compatibility rather than fixed convolutional kernels. Although GATs have been used for point-cloud classification and segmentation, their potential to reconstruct missing LiDAR beams remains largely unexplored. This gap motivates the present work, where GAT is leveraged directly for recovering missing vertical channels in a way that preserves the structural patterns of 3D LiDAR scans.

\subsection{ GAT for Local Feature Aggregation }

\begin{figure}[t]
\centering
\includegraphics[width=0.95\linewidth]{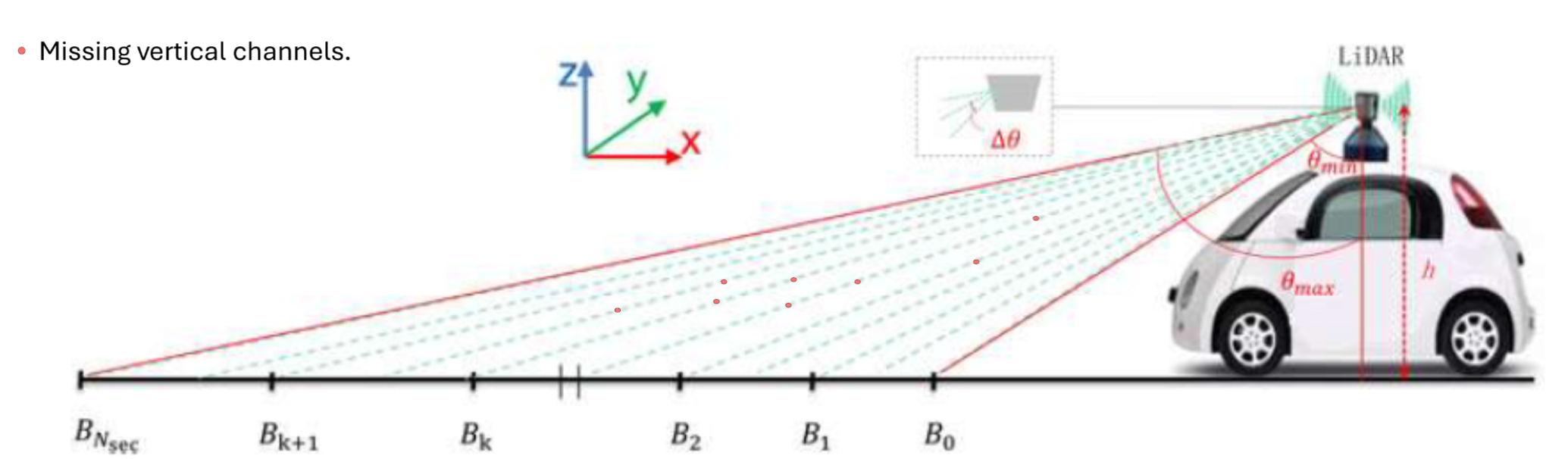}
\caption{LiDAR scanning geometry illustrating the vertical field of view and beam structure, adapted from~\cite{ref15}. Red dots indicate missing points caused by dropped vertical channels.}
\label{fig:Fig01}
\end{figure}

Modern spinning LiDAR sensors acquire a 3D representation of the environment by emitting a set of vertically stacked laser beams while rotating horizontally. Each beam corresponds to a fixed elevation angle, and the collection of these beams forms the vertical channel structure of the sensor. Figure~\ref{fig:Fig01} illustrates this geometry, where the angles $\theta_{\min}$ and $\theta_{\max}$ define the vertical field of view, and the spacing $\Delta\theta$ determines the vertical resolution. Because each beam samples an entire arc of points as the LiDAR rotates, missing a single vertical channel results in the loss of a complete slice of points across the scan.

Each LiDAR return is defined in spherical coordinates by range $r$, elevation angle $\theta$, and azimuth angle $\phi$. These values are converted into Cartesian coordinates through the standard transformation:

\begin{equation}
\begin{aligned}
x &= r \cos\theta \cos\phi,\\
y &= r \cos\theta \sin\phi,\\
z &= r \sin\theta.
\end{aligned}
\label{eq:lidar_coords}
\end{equation}

The elevation angle $\theta$ plays a direct role in determining the vertical coordinate $z$. Consequently, when a vertical channel is dropped—whether due to dust, snow, partial occlusion, or hardware degradation—the corresponding $\theta$ values become unavailable. This leads to systematic gaps in the $z$-dimension, creating vertical discontinuities that degrade downstream perception tasks. Prior studies have shown that the integrity of the vertical channel structure is crucial for reliable object detection and localization~\cite{ref13,ref14}. Similarly, empirical analyses indicate that reductions in vertical resolution significantly impair obstacle detection and distance estimation, particularly for near-field and tall objects~\cite{ref19}.

The annotated points in Figure~\ref{fig:Fig01} highlight where missing vertical channels produce missing returns along the beam trajectory. Recovering these missing $z$-values becomes essential for maintaining geometric consistency in the point cloud and preserving the reliability of autonomous driving perception systems.

\subsection{GAT Operation for Local Neighborhood Reasoning}
\label{subsec:GAT}

GAT extend message–passing graph neural networks by learning
data–dependent weights for each edge instead of using fixed, degree–based normalization
\cite{ref1}. Given a graph $\mathcal{G} = (\mathcal{V}, \mathcal{E})$ with node features
$\mathbf{h}_i \in \mathbb{R}^{F}$ for $i \in \mathcal{V}$, a GAT layer first applies a shared
linear projection
\begin{equation}
\mathbf{h}'_i = \mathbf{W}\,\mathbf{h}_i,
\label{eq:gat_linear}
\end{equation}
where $\mathbf{W} \in \mathbb{R}^{F' \times F}$ is a learnable weight matrix and
$F'$ is the hidden dimension.

For each edge $(i,j) \in \mathcal{E}$, an attention score $e_{ij}$ is computed by
\begin{equation}
e_{ij} = \text{LeakyReLU}\!\big( \mathbf{a}^{\top}
        [\mathbf{Wh}_i \,\Vert\, \mathbf{Wh}_j] \big),
\label{eq:gat_score}
\end{equation}
where $\mathbf{a} \in \mathbb{R}^{2F'}$ is a learnable vector and
$[\cdot \Vert \cdot]$ denotes concatenation. The score is then normalized over
the neighborhood $\mathcal{N}(i)$ using a masked softmax:
\begin{equation}
\alpha_{ij} =
\frac{\exp(e_{ij})}{\sum_{k \in \mathcal{N}(i)} \exp(e_{ik})},
\label{eq:gat_alpha}
\end{equation}
so that $\sum_{j \in \mathcal{N}(i)} \alpha_{ij} = 1$ and $\alpha_{ij}$ measures the
relative importance of node $j$ when updating node $i$.

The new representation of node $i$ is obtained as a weighted aggregation of its neighbors:
\begin{equation}
\mathbf{h}'_i = \sigma\!\left(
    \sum_{j \in \mathcal{N}(i)} \alpha_{ij}\,\mathbf{Wh}_j
\right),
\label{eq:gat_update}
\end{equation}
where $\sigma(\cdot)$ is a nonlinearity (e.g., ELU or LeakyReLU). To improve stability
and expressiveness, GAT commonly uses $K$ independent attention heads
\cite{ref1,ref18}. Each head $k$ computes its own coefficients $\alpha_{ij}^{(k)}$
and embeddings $\mathbf{h}_i^{\prime(k)}$ as in
\eqref{eq:gat_score}–\eqref{eq:gat_update}; the outputs are then concatenated:
\begin{equation}
\mathbf{h}'_i = \big\Vert_{k=1}^{K} \mathbf{h}_i^{\prime(k)}.
\label{eq:gat_multihead}
\end{equation}

\begin{figure}[ht]
    \centering
    \includegraphics[width=1.0\linewidth]{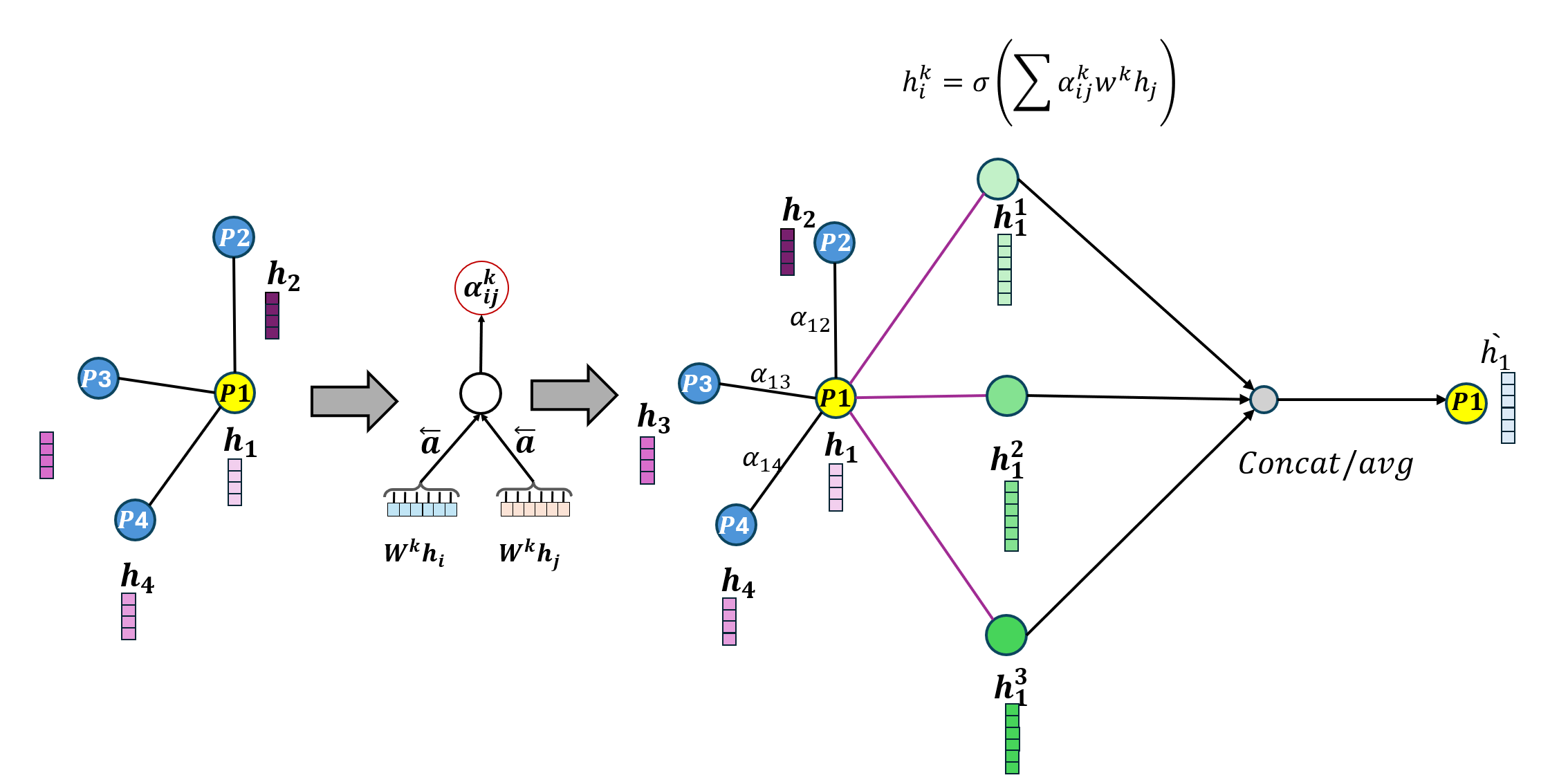}
    \caption{Illustration of a single GAT layer showing linear projection, 
attention computation, multi-head aggregation, and feature update for 
four nodes with three attention heads.}
    \label{fig:GAT_toy}
\end{figure}

Figure~\ref{fig:GAT_toy} illustrates these operations on a example with four
points $P_1,\ldots,P_4$ and three attention heads ($K=3$). Each point $P_i$ is
associated with a feature vector $\mathbf{h}_i$. The features are first projected
by \eqref{eq:gat_linear}, attention scores are computed on the edges, and
head–specific neighbor messages are aggregated according to
\eqref{eq:gat_alpha}–\eqref{eq:gat_update}. The three head outputs for the
central point $P_1$ are finally concatenated as in \eqref{eq:gat_multihead} to
form its updated embedding. This compact example mirrors the same operations
used later in our LiDAR reconstruction model.

\subsection*{C. Multi-Layer GAT Architecture for Vertical Beam Recovery}

The beam-indexed node representation and single-layer attention mechanism described in the previous subsection form the foundation of our recovery model. However, restoring missing vertical channels often requires information beyond a node’s immediate neighborhood—especially when several consecutive beams are absent. To capture broader structural context, we adopt a compact multi-layer GAT architecture where the number of layers is denoted by \(L\).

Prior work has shown that stacking attention layers can strengthen geometric reasoning.~\cite{ref20} showed that multi-head attention benefits from layered aggregation as it mixes information more effectively across the graph. Recent surveys on point-cloud GNNs~\cite{ref17,ref18} further support the use of shallow GAT stacks for tasks involving spatial interpolation and geometry completion.

Given the initial beam-aware features \(\mathbf{h}_i^{(0)}\), each GAT layer expands the receptive field by aggregating information from the local \(k\)NN graph. The update rule for a general \(L\)-layer architecture is:
\begin{equation}
    \mathbf{h}_i^{(l+1)} = \mathrm{GAT}^{(l)}\big(\mathbf{h}_i^{(l)},\, \mathcal{N}(i)\big), \qquad l=0,1,\dots,L-1.
\end{equation}
Feature-wise input normalization is applied prior to graph attention processing to improve numerical stability during training.

After the final layer \(L\), the resulting embedding \(\mathbf{h}_i^{(L)}\) is passed through a lightweight regression head that predicts the corrected vertical coordinate:
\begin{equation}
    \hat{z}_i = \mathbf{W}_2\,\sigma\!\left(\mathbf{W}_1 \mathbf{h}_i^{(L)} + \mathbf{b}_1\right) + b_2,
\end{equation}

where \(\sigma(\cdot)\) is a LeakyReLU activation. This head is specifically optimized for elevation recovery, while the original \((x,y)\) and beam index remain unchanged.

\begin{figure}[ht]
    \centering
    \includegraphics[width=1.0\linewidth]{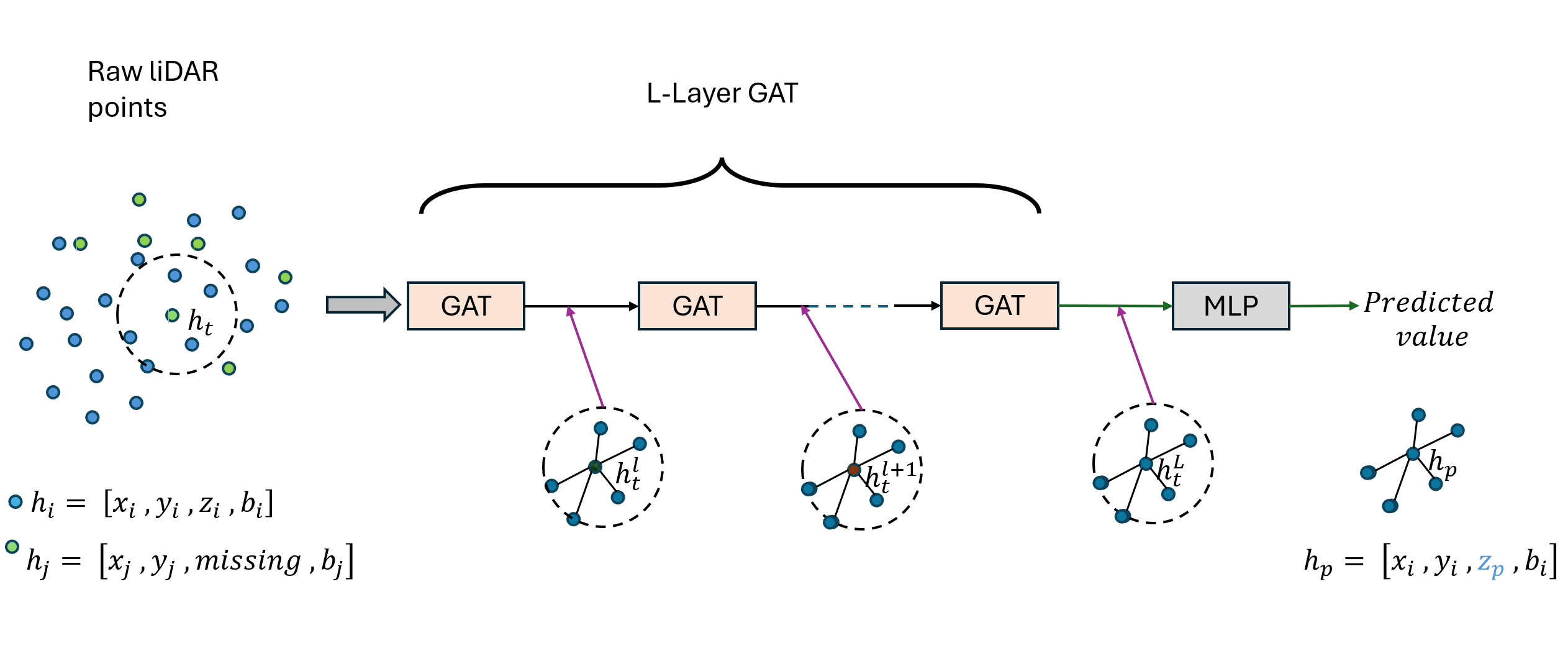}
    \caption{A general \(L\)-layer GAT pipeline for reconstructing missing vertical LiDAR beams using a $k$NN graph. Each layer consists of a GAT block followed by normalization and residual connections.}
    \label{fig:architecture}
\end{figure}

Using \(L\) stacked attention layers enables the model to gather information from progressively wider regions of the LiDAR scan as shown in Figure~\ref{fig:architecture} . This is particularly important for vertical-beam dropout, where restoring the missing points depends on understanding the surrounding elevation structure encoded along the beam direction.

\section{Experimental Setup}

\subsection{Dataset}

We evaluate the proposed method on the KITTI raw dataset, city category, recorded from a vehicle equipped with a Velodyne HDL--64E LiDAR sensor and multiple cameras \cite{ref21}.  
In this work we use sequence contains 1065 synchronized frames (approximately 1.76 minutes).  
Each LiDAR scan provides on the order of $1.2\times 10^5$ 3D points with spatial extents of $x\in[-79.93,\,71.77]$, $y\in[-79.02,\,78.95]$, and $z\in[-9.50,\,2.91]$, and is paired with a front RGB image of resolution $1392 \times 512$.

\begin{figure}[ht]
    \centering
   
   \includegraphics[width=\linewidth]{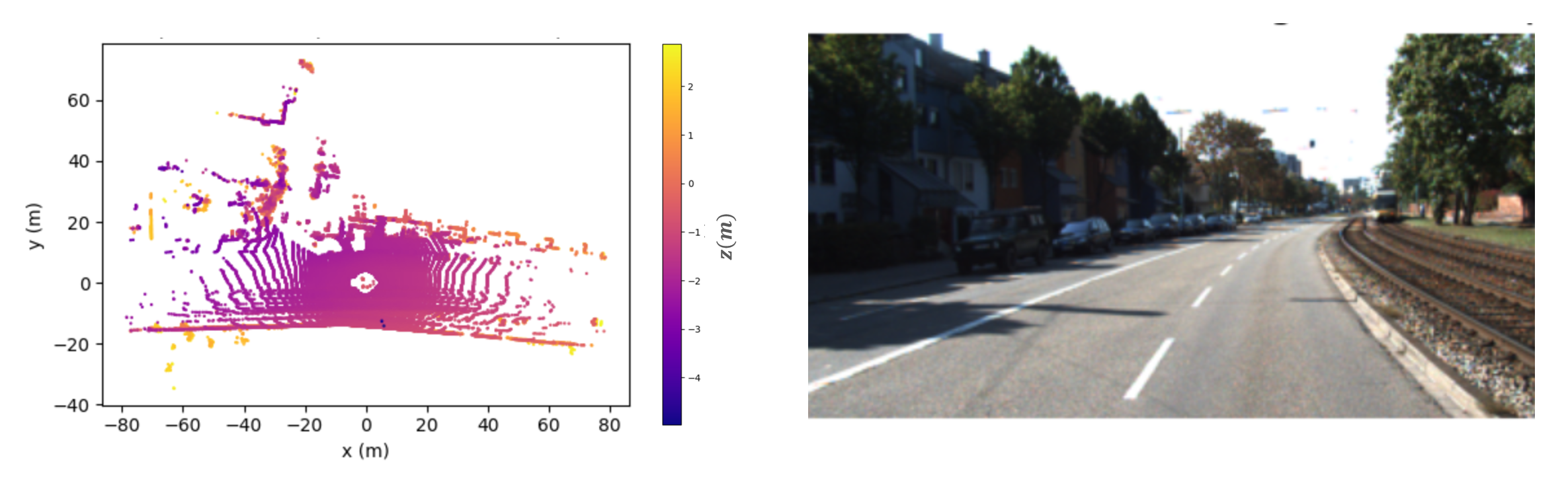}
        \vspace{0.1cm}
        \caption{KITTI sample from sequence \texttt{2011\_09\_29\_drive\_0071}: (a) top-down LiDAR point cloud, (b) corresponding front-view RGB image.}
    \label{fig:Fig4}
\end{figure}

For illustration, Fig.~\ref{fig:Fig4} shows one frame from this sequence: a top-down view of the LiDAR point cloud and the corresponding camera image.

\subsection{Point Cloud Sampling and Graph Construction}

For each frame, we load the raw Velodyne LiDAR scan and retain the 3D coordinates and reflectance,
\begin{equation}
\mathbf{p}_i = [\,x_i,\,y_i,\,z_i,\,b_i\,] \in \mathbb{R}^4 .
\end{equation}
Standard spatial filtering is applied to remove returns outside the valid sensing range, and a uniform random subset of points is kept to obtain a fixed-size input representation. Frames containing an insufficient number of valid points are excluded.

A $k$-nearest-neighbor graph is then constructed in Euclidean $(x,y,z)$ space. Each point becomes a node, and edges connect it to its local geometric neighborhood:
\[
  \mathcal{G} = (\mathcal{V}, \mathcal{E}), \qquad
\mathcal{V} = \{1,\ldots,|\mathcal{V}|\}, \quad
\mathcal{E} = \{(i,j)\mid j \in \mathcal{N}(i)\}. 
\]

The initial node descriptor used by the GAT layers is simply $\mathbf{h}_i^{(0)} = \mathbf{p}_i$, allowing the network to propagate information based solely on spatial position and reflectance along the constructed graph.

\subsection{Implementation Details}

All models are implemented in Python using PyTorch and PyTorch Geometric.  
We rely on standard scientific libraries for data handling and preprocessing.  
Experiments are run on a single GPU in the Kaggle environment, with CUDA enabled. Unless otherwise stated, we use $k=10$ neighbors, a hidden size of 256, 8 attention heads in each GAT layer, a dropout rate of 0.2, and train for up to 100 epochs with early stopping.
\section{Results and Discussion}

We evaluate the proposed graph-attention reconstruction model on a sequence of $50\%$ of LiDAR frames from the KITTI dataset, where every fourth vertical beam is removed to simulate realistic dropout patterns. Our method performs reconstruction directly at the point level without relying on RGB images, voxel discretization, or multi-frame temporal context. Unless otherwise stated, the reported results correspond to the configuration that achieved the highest reconstruction fidelity across the entire sequence. This setup reflects the full capability of our approach and is used as the basis for comparison with existing work. Additional experiments are also presented to illustrate the accuracy--runtime trade-offs when the model is configured for faster inference.
Figure~5 presents a qualitative top--down comparison between the original and reconstructed point clouds for the selected representative frame. Despite the heavy vertical beam dropout, the reconstructed points closely follow the true scan geometry, particularly along the road surface and surrounding structures. The corresponding error distribution is further quantified in Fig.~6 using the cumulative distribution of per-point Euclidean error. The curve shows that a large fraction of the reconstructed points fall within a small spatial deviation from the ground truth, confirming the high geometric fidelity of the proposed method. Quantitatively, the summary results averaged over selected frames are reported in Table~I, where the proposed GAT model achieves a low RMSE$_{XYZ}$ of 0.070\,m, MAE$_z$ of 0.062\,m, and an accuracy of 0.866 within a 10\,cm threshold. A comparison with recent LiDAR reconstruction and sparse-beam recovery approaches is provided in Table~II, demonstrating that our method achieves competitive or superior accuracy without relying on RGB images or temporal multi-frame information.
\begin{figure}[ht]
    \centering
    \includegraphics[width=0.90\columnwidth]{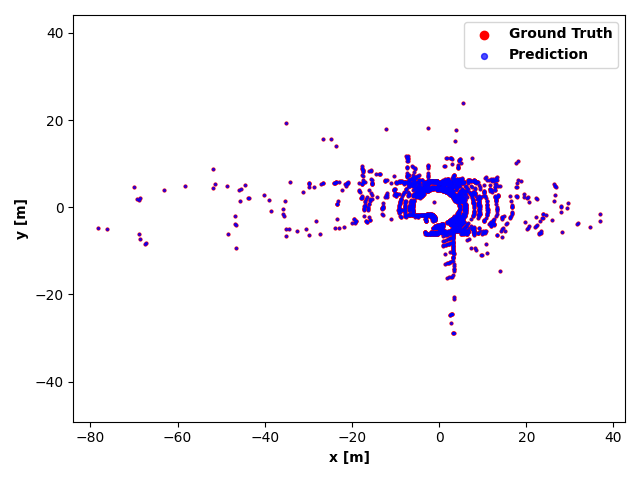}
   \caption{Top--down ($x$--$y$) comparison of ground-truth and reconstructed LiDAR points for a representative frame. Predicted points align closely with the original geometry.}

    \label{fig:topdown_comparison}
\end{figure}
\begin{figure}[ht]
    \centering
    \includegraphics[width=0.90\columnwidth]{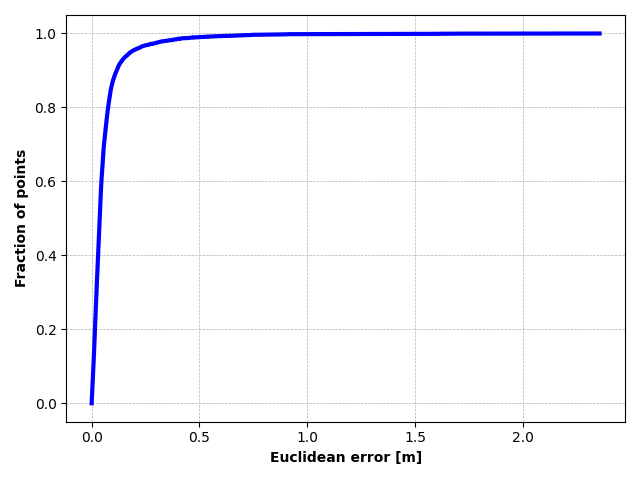}
    \caption{Cumulative distribution of per-point reconstruction error for the representative frame. Most points fall within a small Euclidean deviation from the ground truth, indicating high geometric fidelity under vertical beam dropout.}

    \label{fig:error_cdf}
\end{figure}
In addition to the qualitative analysis, we report the quantitative reconstruction performance of the proposed method averaged across all selected frames. The evaluation is performed exclusively on the masked (dropped) LiDAR channels to avoid inflating the results with observed points. Table~I summarizes the overall accuracy, vertical reconstruction error, geometric consistency, and runtime of the proposed GAT-based model. These results reflect the full-capacity configuration used throughout the comparison with existing methods.

\begin{table}[t]
\centering


\setlength{\tabcolsep}{20pt}  
\renewcommand{\arraystretch}{1.3}

\caption{Average reconstruction performance of the proposed GAT model over selected LiDAR frames.}
\label{tab:main_results}

\begin{tabular}{l r}
\hline
\textbf{Metric} & \textbf{Value} \\
\hline
RMSE$_{XYZ}$ (m)                   & 0.0674 ± 1.13\%\\
RMSE$_z$ (m)                       & 0.1167 ± 1.95\% \\
MAE$_z$ (m)                        & 0.05791 ± 0.76\\
Accuracy                           & 0.8798 ± 2.42\%\\
Chamfer Distance                   & 0.1105 ± 1.45\%\\
Runtime (s/frame)                  & 14.65  ± 1.22\%\\

\hline
\end{tabular}
\end{table}

\begin{figure}[ht]
    \centering
    \includegraphics[width=\columnwidth]{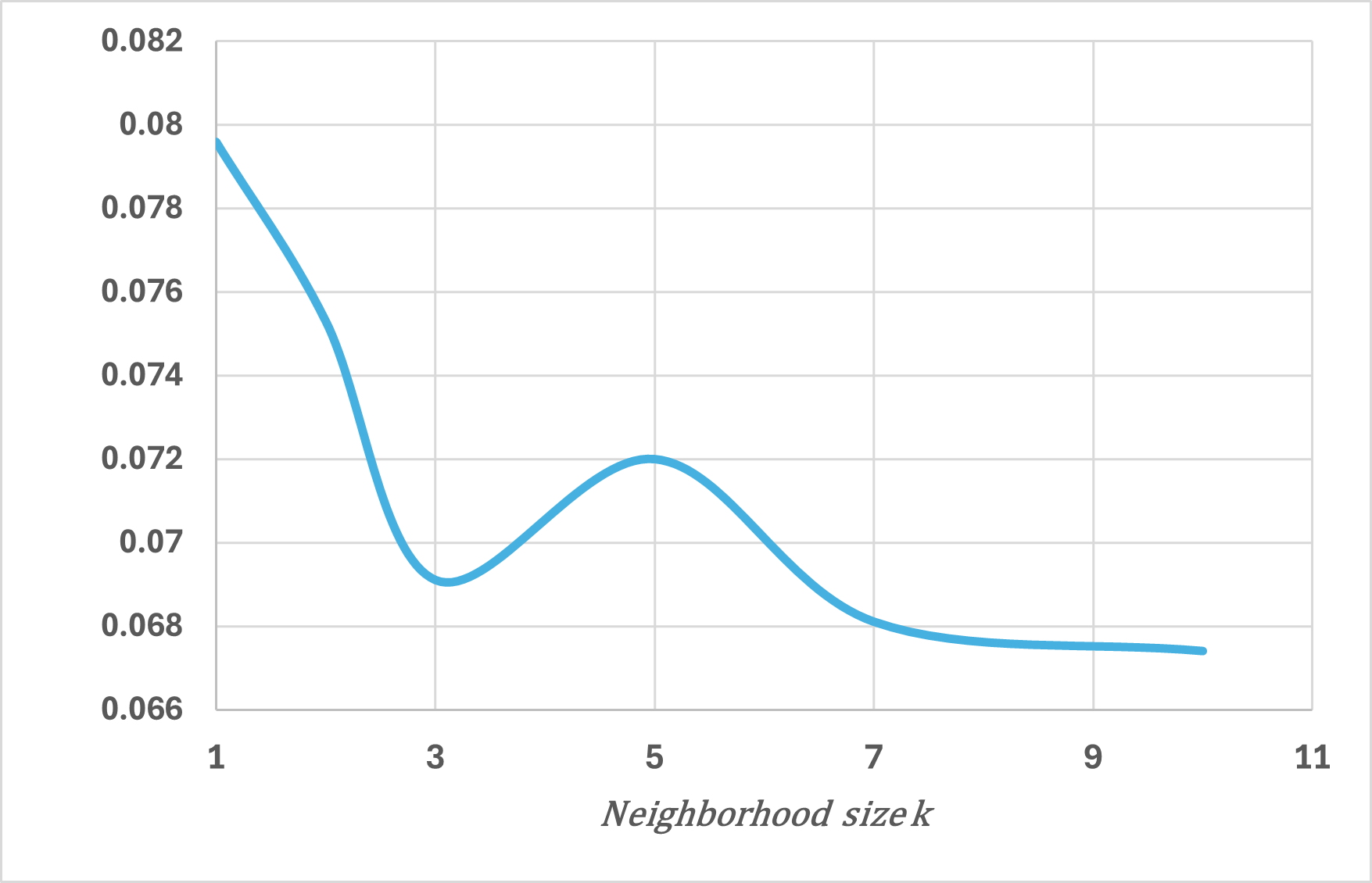}
    \caption{Effect of neighborhood size \(k\) on reconstruction accuracy measured by RMSE$_{XYZ}$. The proposed method maintains stable reconstruction quality across a wide range of neighborhood sizes, with optimal performance observed at larger \(k\).}
    \label{fig:rmse_vs_k}
\end{figure}

\begin{figure}[t]
    \centering
    \includegraphics[width=\columnwidth]{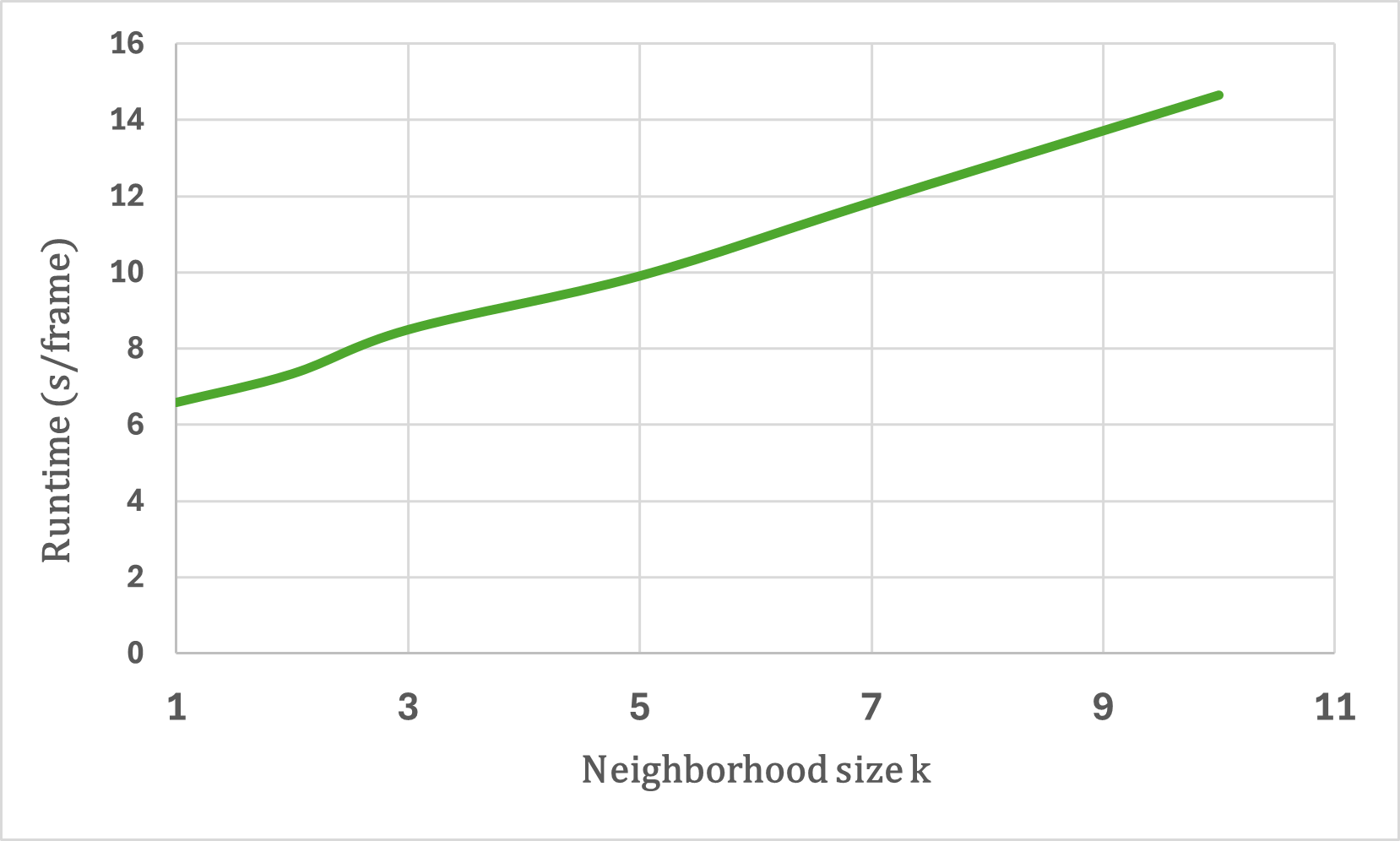}
    \caption{Runtime per frame as a function of neighborhood size \(k\). The computational cost increases monotonically with larger neighborhood sizes due to the higher graph connectivity.}
    \label{fig:time_vs_k}
\end{figure}

To evaluate the flexibility of the proposed LiDAR-only reconstruction framework, we analyze the effect of varying the neighborhood size \(k\) on both reconstruction accuracy and computational cost. Figure~\ref{fig:rmse_vs_k} illustrates the variation of RMSE$_{XYZ}$ with respect to \(k\), while Fig.~\ref{fig:time_vs_k} reports the corresponding runtime per frame. As shown in Fig.~\ref{fig:rmse_vs_k}, the reconstruction quality remains stable across a wide range of neighborhood sizes, indicating that the proposed method is robust even when using relatively small local neighborhoods. At the same time, Fig.~\ref{fig:time_vs_k} shows that the runtime increases steadily with larger \(k\), confirming the expected trade-off between reconstruction accuracy and computational complexity. These results demonstrate that the proposed approach can be efficiently tuned to balance reconstruction quality and processing speed depending on application requirements.

\section{Conclusion}

In this work, we presented a GAT-based framework for reconstructing missing vertical LiDAR beams using only point cloud geometry. By representing each frame as a graph and learning adaptive attention weights among neighboring points, the proposed approach enables accurate recovery of elevation information without requiring RGB images, voxelization, or temporal fusion. Experimental evaluation on KITTI frames demonstrated stable reconstruction accuracy and consistent performance under different neighborhood sizes, while maintaining reasonable runtime.

Unlike many existing methods that rely on RGB or multi-modal fusion, the proposed framework operates directly on raw LiDAR geometry and remains effective under severe vertical beam dropout. This allows reliable reconstruction even in scenarios where camera data is missing, misaligned, or degraded.

\bibliographystyle{IEEEtran}
\bibliography{ref}

@article{ref1,
  title={Graph attention networks},
  author={Velickovic, Petar and Cucurull, Guillem and Casanova, Arantxa and Romero, Adriana and Lio, Pietro and Bengio, Yoshua and others},
  journal={stat},
  volume={1050},
  number={20},
  pages={10--48550},
  year={2017}
}

@inproceedings{ref3,
  title={GLiDR: Topologically regularized graph generative network for sparse LiDAR point clouds},
  author={Kumar, Prashant and Bhat, Kshitij Madhav and Nadkarni, Vedang Bhupesh Shenvi and Kalra, Prem},
  booktitle={Proceedings of the IEEE/CVF Conference on Computer Vision and Pattern Recognition},
  pages={15152--15161},
  year={2024}
}

@inproceedings{ref4,
  title={Sparse points to dense clouds: Enhancing 3D detection with limited LiDAR data},
  author={Kumar, Aakash and Chen, Chen and Mian, Ajmal and Lobo, Neils and Shah, Mubarak},
  booktitle={2024 IEEE/RSJ International Conference on Intelligent Robots and Systems (IROS)},
  pages={12963--12970},
  year={2024},
  organization={IEEE}
}

@article{ref9,
  title={A survey on deep-learning-based lidar 3d object detection for autonomous driving},
  author={Alaba, Simegnew Yihunie and Ball, John E},
  journal={Sensors},
  volume={22},
  number={24},
  pages={9577},
  year={2022},
  publisher={MDPI}
}

@article{ref10,
  title={Cylindrical and asymmetrical 3d convolution networks for lidar-based perception},
  author={Zhu, Xinge and Zhou, Hui and Wang, Tai and Hong, Fangzhou and Li, Wei and Ma, Yuexin and Li, Hongsheng and Yang, Ruigang and Lin, Dahua},
  journal={IEEE Transactions on Pattern Analysis and Machine Intelligence},
  volume={44},
  number={10},
  pages={6807--6822},
  year={2021},
  publisher={IEEE}
}

@article{ref11,
  title={No more potentially dynamic objects: Static point cloud map generation based on 3d object detection and ground projection},
  author={Woo, Soojin and Jung, Donghwi and Kim, Seong-Woo},
  journal={arXiv preprint arXiv:2407.01073},
  year={2024}
}

@article{ref12,
  title={Real time object detection using LiDAR and camera fusion for autonomous driving},
  author={Liu, Haibin and Wu, Chao and Wang, Huanjie},
  journal={Scientific Reports},
  volume={13},
  number={1},
  pages={8056},
  year={2023},
  publisher={Nature Publishing Group UK London}
}

@article{ref13,
  title={Optimizing roadside LiDAR beam distribution to enhance vehicle detection performance considering dynamic vehicle occlusion effects},
  author={He, Yongjiang and Suo, Dajiang and Cao, Peng and Liu, Xiaobo},
  journal={Transportation Research Part C: Emerging Technologies},
  volume={179},
  pages={105268},
  year={2025},
  publisher={Elsevier}
}

@article{ref14,
  title={3D-VDNet: Exploiting the vertical distribution characteristics of point clouds for 3D object detection and augmentation},
  author={Xiao, Weiping and Li, Xiaomao and Liu, Chang and Gao, Jiantao and Luo, Jun and Peng, Yan and Zhou, Yang},
  journal={Image and Vision Computing},
  volume={127},
  pages={104557},
  year={2022},
  publisher={Elsevier}
}

@article{ref15,
  title={Ground-distance segmentation of 3D LiDAR point cloud toward autonomous driving},
  author={Wu, Jian and Yang, Qingxiong},
  journal={APSIPA Transactions on Signal and Information Processing},
  volume={9},
  pages={e24},
  year={2020},
  publisher={Cambridge University Press}
}

@article{ref16,
  title={DenseSphere: Multimodal 3D object detection under a sparse point cloud based on spherical coordinate},
  author={Jung, Jong Won and Yoon, Jae Hyun and Yoo, Seok Bong},
  journal={Expert Systems with Applications},
  volume={251},
  pages={124053},
  year={2024},
  publisher={Elsevier}
}

@article{ref17,
  title={Graph neural networks in point clouds: A survey},
  author={Li, Dilong and Lu, Chenghui and Chen, Ziyi and Guan, Jianlong and Zhao, Jing and Du, Jixiang},
  journal={Remote Sensing},
  volume={16},
  number={14},
  pages={2518},
  year={2024},
  publisher={MDPI}
}

@article{ref18,
  title={Sparse graph attention networks},
  author={Ye, Yang and Ji, Shihao},
  journal={IEEE Transactions on Knowledge and Data Engineering},
  volume={35},
  number={1},
  pages={905--916},
  year={2021},
  publisher={IEEE}
}

@article{ref19,
  title={Impact of LiDAR Vertical Resolution on Object Detection Performance},
  author={Aksamovic, Arnes and Suba{\v{s}}i{\'c}, Mirsad},
  journal={Sensors},
  volume={20},
  number={3},
  pages={826},
  year={2020},
  publisher={MDPI}
}

@article{ref20,
  title={How attentive are graph attention networks?},
  author={Brody, Shaked and Alon, Uri and Yahav, Eran},
  journal={arXiv preprint arXiv:2105.14491},
  year={2021}
}

@article{ref21,
  author = {Andreas Geiger and Philip Lenz and Christoph Stiller and Raquel Urtasun},
  title = {Vision meets Robotics: The KITTI Dataset},
  journal = {International Journal of Robotics Research (IJRR)},
  year = {2013}
}

\end{document}